\def\year{2020}\relax
\documentclass[letterpaper]{article}
\usepackage{aaai20}
\usepackage{times}
\usepackage{helvet}
\usepackage{courier}

\usepackage{url}            % simple URL typesetting
\usepackage{booktabs}       % professional-quality tables
\usepackage{amsfonts}       % blackboard math symbols
\usepackage{amsmath}
\usepackage{bm}
\usepackage{todo}
\usepackage{nicefrac}       % compact symbols for 1/2, etc.
\usepackage{microtype}      % microtypography
\usepackage{algorithm}
\usepackage{mathabx}
\usepackage[noend]{algpseudocode}
\usepackage{verbatim}
\usepackage{breqn}

\usepackage{microtype}
\usepackage{graphicx}
\usepackage{subfigure}
\usepackage{booktabs}

\newcommand{\methodname}{Maximize Overall Diversity}
\newcommand{\methodabbrev}{MOD}
\newcommand{\methodtrain}{MOD-in}

\newcommand{\methodnc}{NegCorr}
\newcommand{\methodnormal}{DeepEns}
\newcommand{\methodnormaladv}{DeepEns+AT}
\newcommand{\methodadv}{MOD-Adv}
\newcommand{\idabbrev}{MOD-R}
\newtheorem{theorem}{Theorem}

\newcommand{\citet}[1]{\citeauthor{#1} \shortcite{#1}} \newcommand{\citep}{\cite} 

\begin{document}
\title{Maximizing Overall Diversity for Improved Uncertainty Estimates 
in Deep Ensembles}
\author{
Siddhartha Jain,\textsuperscript{*\rm 1}
Ge Liu,\textsuperscript{*\rm 1}
Jonas Mueller,\textsuperscript{\rm 2}
David Gifford,\textsuperscript{\rm 1}\\
\textsuperscript{*}The authors contribute equally,
\textsuperscript{\rm 1}CSAIL,MIT,
\textsuperscript{\rm 2}Amazon Web Services\\
sj1@mit.edu, geliu@mit.edu, jonasmue@amazon.com,
gifford@mit.edu
}
\maketitle
\begin{abstract}
\begin{quote}
The inaccuracy of neural network models on inputs that do not stem from the distribution underlying the training data is problematic and at times unrecognized. 
Uncertainty estimates of model predictions are often based on the variation in predictions produced by a diverse ensemble of models applied to the same input.
Here we describe \methodname{} (\methodabbrev{}), an approach to improve ensemble-based uncertainty estimates by encouraging larger overall diversity in ensemble predictions across all possible inputs. % We also explore variations of \methodabbrev{} utilizing adversarial techniques (\methodadv{}) and data density estimation (\idabbrev{}). 
We apply \methodabbrev{} to regression tasks including 38 Protein-DNA binding datasets, 9 UCI datasets, and the IMDB-Wiki image dataset. We also explore variants that utilize adversarial training techniques and data density estimation. 
For out-of-distribution test examples, \methodabbrev{} significantly improves predictive performance and uncertainty calibration 
 without sacrificing performance on test data drawn from same distribution as the training data.
 We also find that in Bayesian optimization tasks, the performance of UCB acquisition is improved via \methodabbrev{} uncertainty estimates. 
\end{quote}
\end{abstract}

\section{Introduction}
\label{intro}
Model ensembling provides a simple, yet extremely effective technique for improving the predictive performance of arbitrary supervised learners each trained with empirical risk minimization (ERM) \cite{bootstrap,Brown04}.  Often, ensembles are utilized not only to improve predictions on test examples stemming from the same underlying distribution as the training data, but also to provide estimates of model uncertainty when learners are presented with out-of-distribution (OOD) examples that may look different than the data encountered during training \cite{Lakshminarayanan17,bootstrapdqn}. 
The widespread success of ensembles crucially relies on the variance-reduction produced by aggregating predictions that are statistically prone to different types of individual errors \cite{Kuncheva03}. 
Thus, prediction improvements are best realized by using a large ensemble with many base models, and a large ensemble is also typically employed to produce stable distributional estimates of model uncertainty \cite{bootstrap,Papadopoulos01}.

\begin{figure*}[ht!]
\centering
  \includegraphics[width=1.\linewidth,trim=5.7cm 0cm 5.2cm 0cm, clip=true]{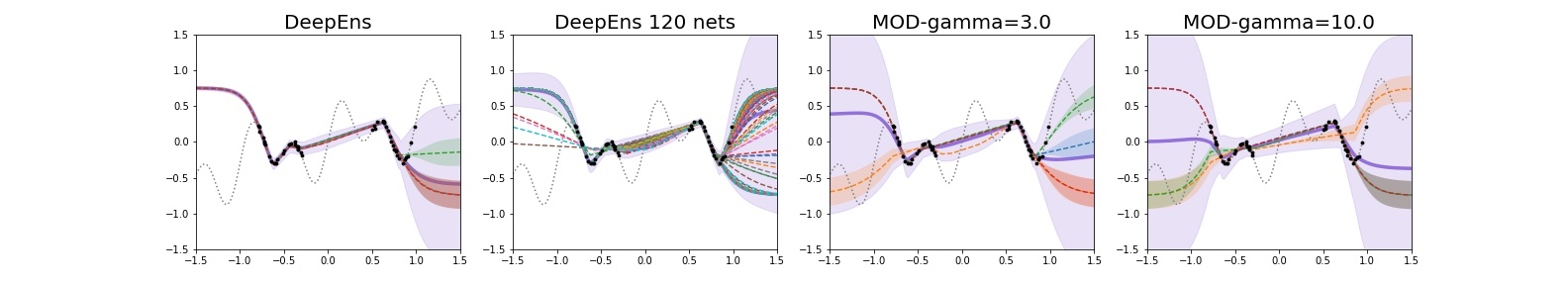}

\caption{Regression on synthesized data with $95\%$ confidence intervals (CI). The training examples are depicted as black dots and the ground-truth function as a grey dotted line.
The predicted conditional mean and CI from individual networks are drawn in colored dashed lines/bands and the overall ensemble conditional mean/CI are depicted via the smooth purple line/band.}
\label{fig:toy}
\end{figure*}

Practical applications of massive neural networks (NN) are commonly limited to small ensembles because of the unwieldy nature of these models \cite{bootstrapdqn,darkknowledge,activelearning}.
Although supervised learning performance may be enhanced by an ensemble comprised of only a few ERM-trained models, the resulting ensemble-based uncertainty estimates can exhibit excessive sampling variability in low-density regions of the underlying training distribution. Consider the example of an ensemble comprised of five models whose predictions just might agree at points far from the training data by chance. 
Figure \ref{fig:toy} depicts an example of this phenomenon, which we refer to as \emph{uncertainty collapse}, since the resulting ensemble-based uncertainty estimates would indicate these predictions are of high-confidence despite not being supported by any nearby training datapoints.

Unreliable uncertainty estimates are highly undesirable in applications where future input queries may not stem from the same distribution.   A shift in input distribution can be caused by sampling bias, covariate shift, and the adaptive experimentation that occurs in bandits, Bayesian optimization (BO), and reinforcement learning (RL) contexts.
Here, we propose \methodname{} (\methodabbrev{}), a technique to stabilize OOD model uncertainty estimates produced by an ensemble of arbitrary neural networks. 
The core idea is to consider \emph{all} possible inputs and encourage as much overall diversity in the corresponding model ensemble outputs as can be tolerated without diminishing the ensemble's predictive performance.  
\methodabbrev{} utilizes an auxiliary loss function and data-augmentation strategy that is easily integrated into any existing training procedure.

\section{Related Work} 

NN ensembles have been previously demonstrated to produce useful uncertainty estimates for sequential experimentation applications in Bayesian optimization and reinforcement learning \cite{Papadopoulos01,Lakshminarayanan17,deepbandit}.
% ,bootstrapdqn,Chen17}. 
Proposed methods to improve ensembles include adversarial training to enforce smoothness~\cite{Lakshminarayanan17}, and maximizing ensemble output diversity over the training data \cite{Brown04}. Recent work has proposed regularizers based on augmented out-of-distribution examples, but is primarily specific to classification tasks and non-trivially requires auxiliary generators of OOD examples 
%use augmentation set from other classes or auxiliary generators that produce points that are close to training distribution
\cite{Lee18} or existing examples from other classes \cite{vyas2018out}. Another line of related work solely aims  at producing better out-of-distribution detectors~\cite{liang2017enhancing,choi2018generative,ren2019likelihood}.

Our work seeks to improve uncertainty estimates in regression settings, where OOD data can stem from an arbitrary unknown distribution, and robust prediction on OOD data is desired rather than just detection of OOD examples. We propose a simple technique to regularize ensemble behavior over \emph{all} possible inputs that does not require training of additional generator.
Consideration of all possible inputs has previously been advocated by \cite{dar}, although not in the context of uncertainty estimation. 
\citet{pearce2018uncertainty} propose a regularizer to ensure an ensemble approximates a valid Bayesian posterior, but their methodology is only applicable to homoskedastic noise unlike ours. 
\citet{Hafner18} also aim to control Bayesian NN output-behavior beyond the training distribution, but our methods do not require the Bayesian formulation they impose and can be applied to arbitrary NN ensembles, which are one of the most straightforward methods used for quantifying NN uncertainty \citep{Papadopoulos01,Lakshminarayanan17,deepbandit}.
% Contrary to prior works, we propose a simple technique to regularize ensemble behavior over \emph{all} possible inputs instead of neighborhood of train distribution, which does not require training of additional generator.
% enforce a Bayesian posterior on the ensemble. However they assume that the noise is homoskedastic whereas we allow for heteroskedastic noise. 
\citet{malinin2018predictive} focus on incorporating distributional uncertainty into uncertainty estimates via an additional prior distribution, whereas our focus is on improving model uncertainty in model ensembles.

% While we primarily consider regression settings here, our ideas can be easily adapted to classification by replacing variance terms with entropy terms; a similar variant that relies on an auxiliary generator network to produce augmented samples that do not stem from the training distribution has been recently proposed by \cite{Lee18}.

\section{Methods}
\label{method}

% \subsection{General problem setup}

%to model the probabilistic predictive distribution $p_\theta(y|x)$ over the labels and measure model uncertainty, where $\theta$ are the parameters of the NN. We introduce a new training loss -- maximum overall diversity (MOD) loss which uses augmented data to improve uncertainty estimation on out-of-distribution region.  
%\subsection{Training ensemble of neural networks}

We consider standard regression, assuming  continuous target values are generated via 
${Y = f(X) + \epsilon}$ with $\epsilon \sim N(0, \sigma_X^2)$, such that $\sigma_X$ may heteroscedastically depend on feature values $X$.
Given a limited training dataset $\mathcal{D} = \{x_n, y_n\}^N_{n=1}$, where $x_n \sim P_{in}$ specifies the underlying data distribution from which the  \emph{in-distribution} examples in the training data are sampled, our goal is to learn an  ensemble of $M$ neural networks that accurately models both the underlying function $f(x)$ as well as the uncertainty in ensemble estimates of $f(x)$. 
Of particular concern are scenarios where test examples $x$ may stem from a different distribution $P_{out} \neq P_{in}$, which we refer to as \emph{out-of-distribution} (OOD) examples. As in \cite{Lakshminarayanan17}, each network $m$ (with parameters $\theta_m$) in our NN ensemble outputs both an estimated mean $\mu_m(x)$ to predict $f(x)$ and an  estimated variance $\sigma^2_m(x)$ to predict $\sigma_x^2$, and the per network loss function ${L(\theta_m; x_n, y_n)} = {-\log p_{\theta_m}(y_n|x_n)}$, is chosen as the negative log-likelihood (NLL)
under the Gaussian assumption ${y_n \sim N(\mu_m(x_n), \sigma^2_m(x_n) )}$. 
% $$L(f_m(x),y)=-\log p_{\theta_m}(y_n|x_n)=\frac{\log\sigma^2_{\theta_m}(x_i)}{2}+\frac{(y_i-\mu_{\theta_m}(x_i))^2}{2\sigma^2_{\theta_m}(x_i)}$$
While traditional bagging provides different training data to each ensemble member,  we simply train each NN using the entire dataset, since the randomness of separate NN-initializations and SGD-training suffice to produce comparable performance to bagging of NN models   \cite{Lakshminarayanan17,Lee15,bootstrapdqn}.

Following \cite{Lakshminarayanan17}, we estimate $P_{Y\mid X=x}$ (and NLL with respect to the ensemble) by treating the aggregate ensemble output as a single Gaussian distribution $N(\widebar{\mu}(x), \widebar{\sigma}^2(x))$.
Here, the ensemble-estimate of $f(x)$ is given by  $\widebar{\mu}(x) = \text{mean}\big(\{\mu_m(x)\}_{m=1}^M\big)$, and the uncertainty in the target value is given by $\widebar{\sigma}^2(x) = \sigma_{\text{eps}}^2(x) +\sigma_{\text{mod}}^2(x)$ based on noise-level estimate $\sigma_{\text{eps}}^2(x) = \text{mean}\big(\{\sigma_m^2(x)\}_{m=1}^M\big)$ and model uncertainty estimate  $\sigma_{\text{mod}}^2(x) = \text{variance}\big(\{\mu_m(x)\}_{m=1}^M\big)$.
While we focus on Gaussian likelihoods for simplicity, our proposed methodology is applicable to general parametric conditional distributions.

\subsection{Maximizing Overall Diversity (MOD) }

Assuming $X \in \mathcal{X}, Y \in [-C, C]$ have been scaled to bounded regions, \methodabbrev{} encourages higher ensemble diversity by introducing an auxiliary loss that is computed over augmented data sampled from another distribution $Q_X$. Like $P_{in}$, $Q_X$ is also defined over the input feature space, but  differs from the underlying training data distribution and instead describes OOD examples that could be encountered at test-time.
% to improve uncertainty estimation on out-of-distribution region by increasing diversity, where we define diversity as the variance of ensemble predictions.
%\subsection{Maximizing overall diversity for better out-of-distribution uncertainty}
%Unlike [\cite{}] which uses an extra loss term to maximize the variance of predictions from the ensemble on the training samples only, we want to take into consideration of the model uncertainty on the out-of-distribution test samples by choosing model $f$ via following mixture-objective:
%$$\min_{f\in\mathcal{H}} (1-\gamma)\cdot\mathbb{E}_{P_{in}}[L(f(x),y)]-\gamma\cdot\mathbb{E}_{P_{out}}[Diversity[f(x)]]$$
The underlying population objective we target is
\begin{equation}\label{eq:obj}
\begin{split}
\min_{\theta_1,\dots,\theta_M} &\ L_{in} - \gamma L_{out}
\ \ \ \text{ where } \\
L_{in} &= \frac{1}{M} \sum_{m=1}^M \mathbb{E}_{P_{{in}}} [L(\theta_m, x,y)]\\
L_{out} &= \mathbb{E}_{Q}[\sigma_{\text{mod}}^2(x)]
\end{split}
\end{equation} 

\iffalse
\[
\text{where } \ \  L_{in} = \frac{1}{M} \sum_{m=1}^M \mathbb{E}_{P_{{in}}} [L(\theta_m, x,y)] 
\ \text{ and } \ 
L_{out} = \int_{\mathcal{X}}  \sigma_{\text{mod}}^2(x) q(x) \mathrm{d} x
\]
\fi 
  
with $L$ as the original supervised learning loss function (e.g.\ NLL), and a user-specified penalty $\gamma > 0$. Since NLL entails a proper-scoring rule \cite{Lakshminarayanan17}, minimizing the above objective with a sufficiently small value of $\gamma$ will ensure the ensemble seeks to recover $P_{Y\mid X=x}$ for inputs $x$ that lie in the support of the training distribution $P_{{in}}$ and otherwise output large model uncertainty for OOD $x$ that lie outside this support. 
As it is difficult in most applications to specify how future OOD examples may look, we aim to ensure our ensemble outputs high uncertainty estimates for any possible $P_{out}$ by taking the entire input space into consideration. To account for any possible OOD distribution, we simply pick $Q_X$ as the uniform distribution over $\mathcal{X}$, the bounded region of all possible inputs $x$.
% $q(x) = 1/\text{Vol}(\mathcal{X})$ where $\mathcal{X}$ is the bounded region of all possible inputs $x$ and $q(x)$ denotes the pdf (or pmf) of $Q_X$. 
This choice is motivated by Theorem \ref{thm:uniform} below, which states that the uniform distribution most closely approximates all possible OOD distributions in the minimax sense. 

\begin{theorem} 
The uniform distribution $Q_X$ equals:
$ \displaystyle \arg\min_{Q \in  \mathcal{P}} \ \max_{P_{out} \in \mathcal{P}} \ \text{KL} (P||Q) $
% = \mathcal{U}$ where $\mathcal{U}$ is the uniform distribution.
\label{thm:uniform}
where for discrete $\mathcal{X}$, $\mathcal{P}$ denotes the set of all distributions, and for continuous $\mathcal{X}$, $\mathcal{P}$ is the set of all distributions with density functions that are bounded within some interval $[a, b]$.
\end{theorem}
\textbf{Proof} For the discrete case with $| \mathcal{X}| = N$: let $P_{out}, Q$ have corresponding pmf $p, q$, so $KL(P_{out}||Q)
%= H(P_{out}, Q) - H(P_{out}) 
= \sum_{x \in \mathcal{X}} p(x) \log p(x) - \sum_{x\in \mathcal{X}} p(x) \log q(x)$.
%where $H(\cdot)$ denotes entropy and $H(\cdot,\cdot)$ cross-entropy. 
When $Q$ is the uniform distribution, the worst case $P_{out}$ is one that puts all its mass on a single point $x$, which corresponds to $KL(P_{out}||Q) = \log N$. For any non-uniform $Q'$: there exists $x'$ where $q'(x') < q(x') = 1/N$. Thus for $P'_{out}$ which puts all its mass on $x'$, we have $KL(P'_{out}||Q') > \log N$. % Thus the worst case $KL$ for any non-uniform $Q$ is higher than the worst case $KL$ for uniform $Q$. 
The proof for the continuous case is similar. \hfill\ensuremath{\blacksquare}

In practice, we approximate $L_{in}$ using the average loss over the training data as in ERM, and train each $\theta_m$ with respect to its contribution to this term independently of the others as in bagging. 
To approximate $L_{out}$, we similarly utilize an empirical average based on augmented examples $\{x_j\}_{j=1}^K$ sampled uniformly throughout the feature space $\mathcal{X}$. Uniformly sampling from the input space takes constant time to compute. We expect only a marginal increase in terms of training time since the computation of back-propagation is largely parallelized and thus an increase in minibatch size would only cause an increase in memory consumption rather than computation time. The formal \methodabbrev{} procedure is detailed in Algorithm~\ref{mod}.
We advocate selecting $\gamma$ as the largest value for which estimates of $L_{in}$ (on held-out validation data) do not indicate worse predictive performance.
This strategy naturally favors smaller values of $\gamma$ as the sample size $N$ grows, thus resulting in lower model uncertainty estimates (with $\gamma \rightarrow 0$ as $N \rightarrow \infty$ when $P_{in}$ is supported everywhere and our NN are universal approximators).

We also experiment with an alternative choice of $Q_X$ being the uniform distribution over the finite training data (i.e.\ ${q(x) = 1/N} \ \ \forall x \in \mathcal{D}$ and {= 0} otherwise).
% , where $\mathcal{D}$ is the training set of $N$ examples). 
We call this alternative method \methodtrain{}, and note its similarity to the diversity-encouraging penalty proposed by \cite{Brown04}, which is also measured over the training data. 
Note that \methodabbrev{} in contrast considers $Q_X$ to be uniformly distributed over all possible test inputs rather than only the training examples.
Maximizing diversity solely over the training data may fail to control ensemble behavior at OOD points that do not lie near any training example, and thus fail to prevent uncertainty collapse.

%  \subsection{Reweighting Augmented Samples by distance (\idabbrev{})}
\subsection{Maximizing Reweighted Diversity (\idabbrev{})}
Aiming for high-quality OOD uncertainty estimates, we are mostly concerned with regularizing the ensemble-variance around points located in low density regions of the training data distribution. 
To obtain a simple estimate that intuitively reflects the inverse of the local density of $P_{in}$ at a particular set of feature values, one can compute the feature-distance to the nearest training data points \cite{papernot2018deep}. 
Under this perspective, we want to encourage greater model uncertainty for the lowest density points that lie furthest from the training data. Commonly used covariance kernels for Gaussian Process regressors (e.g.\ radial basis functions) explicitly enforce a high amount of uncertainty on points that lie far from the training data.
As calculating the distance of each point to the entire training set may be undesirably inefficient for large datasets, we only compute the distance of our augmented data to a current minibatch $\mathcal{B}$  during training. Specifically, we use these distances to compute the following:
\begin{equation} \label{eq:id}
    \tilde{w}_b = \frac{\sum_{i=1}^k ||\tilde{x}_b-x^b_i||_2^2}{\max_{b} \sum_{i=1}^k ||\tilde{x}_b-x^b_i||_2^2}
\end{equation}
where $\tilde{w}_b$ are weights for each of the augmented points $\tilde{x}_b$, and $x^b_i  \ (i=1,\dots,k)$ are members of the minibatch $\mathcal{B}$ that are the $k$ nearest neighbors of $\tilde{x}_b$. Throughout this paper, we use $k=5$. 

The $\tilde{w}_b$ are thus inversely related to a crude density estimate of the training distribution $P_{in}$ evaluated at each augmented sample $\tilde{x}_b$.  
Rather than optimizing the loss $L_{out}$ which uniformly weights each augmented sample (as done in Algorithm 1), we can instead form a weighted loss computed over the minibatch of augmented samples as:
$
\sum_{b=1}^{|\mathcal{B}|} \tilde{w}_b \cdot \sigma^2_{\text{mod}}(\widetilde{x}_b) 
$
which should increase the model uncertainty for augmented inputs proportionally to their distance from the training data.
We call this variant of our methodology with augmented input  reweighting \idabbrev{}.

\subsection{Maximizing Overall Diversity with Adversarial Optimization (\methodadv{})}

We also consider another variant of \methodabbrev{} that utilizes adversarial training techniques. Here, we maximize the variance on relatively over-confident points in out-of-distribution regions, which are likely to comprise worst-case $P_{out}$. Specifically, we formulate a maximin optimization for the \methodabbrev{} penalty $\displaystyle \max_{\Theta}\min_{x}\sigma^2_{\text{mod}}(x)$, and thus the full training objective becomes
$\min_{\theta_1,\dots,\theta_M} L_{in} - \gamma\cdot \min_x \sigma^2_{\text{mod}}(x)$.
We call this variant \methodadv{}. In practice, we obtain the augmented points by taking a single gradient step in the direction of lower variance ($\sigma^2_{\text{mod}})$, starting from uniformly sampled points. The extra gradient step can double the computation time  compared to MOD. The full algorithm is given in Algorithm~\ref{mod}. Note that  \methodadv{} is different than the traditional adversarial training in two aspects: first it takes a gradient step with regard to the model uncertainty measurement (the variance of ensemble mean prediction) instead of with regard to the predicted score of another class; second, the adversarial step is taken starting from a uniformly sampled example instead of a training example. We apply  \methodadv{} to only regression tasks with continuous features since it is more natural to apply gradient descent on them.

\begin{algorithm}
\caption{ \ \methodabbrev{} Training Procedure \ (+ Variants)}\label{mod}
\begin{algorithmic}
\State \textbf{Input:} Training data $\mathcal{D} = \{(x_n, y_n)\}_{n=1}^N$,  penalty $\gamma > 0$, batch-size $|\mathcal{B}|$
\State \textbf{Output:} Parameters of ensemble of $M$ neural networks $\theta_1$,...,$\theta_M$
\State Initialize $\theta_1$,...,$\theta_M$ randomly, initialize $w_b = 1$ for $b = 1,\dots,|\mathcal{B}|$
\Repeat 
\State Sample minibatch from training data:$\{(x_b,y_b)\}_{b=1}^{|\mathcal{B}|}$
\State Sample $|\mathcal{B}|$ augmented inputs $\widetilde{x}_1$,..., $\widetilde{x}_B$ uniformly at random from $\mathcal{X}$
\If {\methodadv{}} 
\State $\tilde{x}_b \leftarrow \widetilde{x}_b - \alpha_{adv}\cdot \nabla_{\widetilde{x}_b} \sigma^2_{\text{mod}}(\widetilde{x}_b) \ \ \forall 1 \leq b \leq |\mathcal{B}|$
\EndIf
% \State Compute empirical objective for the ensemble $J(\theta)$
\For{m = 1,\dots, M}
    \If {\idabbrev{}} \ \ $w_b = \tilde{w}_{b}$ \ (defined in equation~(\ref{eq:id}))
    \EndIf 
    \State Update $\theta_m$ via SGD with gradient $ =\displaystyle \frac{1}{|\mathcal{B}|} \nabla_{\theta_m} \bigg[ \sum_{b=1}^{|\mathcal{B}|} L(\theta_m;(x_b,y_b)) - \gamma \sum_{b=1}^{|\mathcal{B}|} w_b \cdot \sigma^2_{\text{mod}}(\widetilde{x}_b)  \bigg] $
\EndFor
\Until{iteration limit reached}
\end{algorithmic}
\end{algorithm}
\setlength\tabcolsep{3.5pt}

\section{Experiments}
\label{result}
%\todo[comment on how we ensure predictive variance > 0]

\subsection{Baseline Methods}
\label{baseline}
Here, we evaluate various alternative strategies for improving model ensembles. All strategies are applied to the same base NN ensemble, which is taken to be the Deep Ensembles  (\methodnormal{}) model of \cite{Lakshminarayanan17} previously described in \textbf{Methods}.

\subsubsection{Deep Ensembles with Adversarial Training (\methodnormaladv{})}

\cite{Lakshminarayanan17} used this strategy to improve their basic \methodnormal{} model.  The idea is to adversarially sample inputs that lie close to the training data but on which the NLL loss is high (assuming they share the same label as their neighboring training example).  Then, we include these adversarial points as augmented data when training the ensemble, which smooths the function learned by the ensemble. 
Starting from training example $x$, we sample augmented datapoint  $x' = x + \delta \text{sign}(\nabla_x L(\theta, x, y))$ with the labels for $x'$ assumed to be the same as that for the corresponding $x$. $L$ here denotes the NLL loss function, and the values for hyperparameter $\delta$ that we search over include 0.05, 0.1, 0.2.

\subsubsection{Negative Correlation (\methodnc{}) } 
This method from ~\cite{liu1999ensemble,Shui18} minimizes the empirical correlation between predictions of different ensemble members over the training data. It adds a penalty to the loss of the form $\sum_m [(\mu_m(x) - \widebar{\mu}(x)) \cdot \sum_{m'\neq m} (\mu_{m'}(x) - \widebar{\mu}(x))]$ where $\mu_m(x)$ is the prediction of the $m$th ensemble member and $\widebar{\mu}(x)$ is the mean ensemble prediction. 
% We refer to this method as \methodnc{}. 
This penalty is weighted by a user-specified penalty $\gamma$, as done in our methodology.

\subsection{Experiment Details}

All experiments were run on Nvidia TitanX 1080 Ti and Nvidia TitanX 2080 Ti GPUs with PyTorch version 1.0. % The image regression task was run only on the 1080 Ti. 
Unless otherwise indicated, all p-values were computed using a single tailed paired t-test per dataset, and the p-values are combined using Fisher's method to produce an overall p-value across all datasets in a task. All hyperparameters -- including learning rate, $\ell_2$-regularization, $\gamma$ for MOD/Negative Correlation, and adversarial training $\delta$ -- were tuned based on validation set NLL. In every regression task, the search for hyperparameter $\gamma$ was over the values 0.01, 0.1, 1, 5, 10, 20, 50. For \methodadv{}, we search for $\delta$ over 0.2,1.0,3.0,5.0 for UCI and 0.1,0.5,1 for the image data.

\subsection{Univariate Regression}
We first consider a one-dimensional regression toy dataset that is similar to the one used by \cite{blundell2015weight}. We generated training data from the function:
$$
y = 0.3x + 0.3 \sin(2\pi x)+0.3 \sin(4\pi x)+\epsilon
$$
$$
\ \ \ \text{ with } \ \epsilon \sim \mathcal{N}(0,0.02)
$$
Here, the training data only contain samples drawn from two limited-size regions. Using the standard NLL loss as well as the auxiliary MOD penalty, we train a deep ensemble with 4 neural networks  of identical architectures  consisting of 1-hidden layer with 50 units, ReLU activation, two sigmoid outputs to estimate the mean and variance of $P_{Y\mid X = x}$, and L2 regularization. To depict the improvement gained by simply adding ensemble members, we also train an ensemble of 120 networks with same architecture. Figure~\ref{fig:toy} shows the predictions and $95\%$ confidence interval of the ensembles. MOD is able to produce more reliable uncertainty estimates on the lefthand regions that lack training data, whereas standard deep ensembles exhibit uncertainty collapse, even with many networks. MOD also properly inflated the predictive uncertainty in the center region where no training data is found. Using a smaller $\gamma=3$ in MOD ensures the ensemble predictive performance remains strong for in-distribution inputs that lie near the training data and the ensemble exhibits adequate levels of certainty around these points.  While the larger $\gamma=10$ value leads to overly conservative uncertainty estimates that are large everywhere, we note the mean of the ensemble predictions remains highly accurate for in-distribution inputs.

\subsection{Protein Binding Microarray Data}
We next study scientific data with discrete features by predicting Protein-DNA binding. This is a collection of 38 different microarray datasets, each of which contains measurements of the binding affinity of a single transcription factor (TF) protein against all possible 8-base DNA sequences \cite{barrera2016survey}. 
We consider each dataset as a separate task with $Y$ taken to be the binding affinity scaled to the interval [0,1] and $X$ the one-hot embedded DNA sequence.  As we ignore reverse-complements, there are $\sim 32,000$ possible values of $X$.

% \iffalse
% consisting of 38 assays \cite{} which test protein binding over DNA sequences \cite{barrera2016survey}. In each assay, the binding affinity (as measured by fluorescence) between a particular DNA-binding protein (transcription factor,TF) and all 8-base DNA sequences are measured as a continuous value with scale ranging from $10^4$ to $10^5$. All possible 8-base sequences consist of 4 nucleotides(TCGA) are tested where the reverse compliments are removed, which gives $\sim 32,000$ pairs of sequence-label pairs. Each sequence is one-hot embedded into $4\times8$ matrix and then vectorized into a 32-dimensional vector which is used as the input. We use all 38 TF families and train models on each dataset to evaluate the overall performance of MOD in both regression task and Bayesian optimization task.
% \fi

\paragraph{Regression}
We trained a small ensemble of 4 neural networks with the same architecture as in the previous experiments.
%(1-hidden layer with 50 units, ReLU activation). 
We consider 2 different OOD test sets, one comprised of the sequences with top 10\% $Y$-values and the other comprised of the sequences with more than 80\% of the position in $X$ being G or C (GC-content). For each OOD set, we use the remainder of the sequences as corresponding in-distribution set. We separate them into extremely small training set (300 examples) and validation set (300 examples), and use the rest as in-distribution test set. We compare \methodabbrev{} along with 3 alternative sampling distribution (\methodtrain{}, \idabbrev{}, and \methodadv{}) against the 3 baselines previously mentioned. We search over 0,1e-3,0.01,0.05,0.1 for $l2$ penalty and 0.01 for learning rate.
% \iffalse
% We held out top 10\% of the data as OOD test set as in Bayesian optimization the true max value distribution is usually not captured when the initial training set is small. 
% We randomly separate the rest into training/validation/in-distribution test set. We experiment with the case where the training set is extremely small (300 examples) and use a small validation set (300 examples) for hyper-parameter tuning. We take the log of affinity measurement and z-score normalize the labels in training set and apply the same normalization on test and validation sets. Then we sigmoid transformed all the labels. 
% \fi 

\begin{table*}[htb!]
\small 
\centering\caption{NLL on OOD/in-distribution test set averaged across 38 TFs over 10 replicate runs(See Appendix Table 1 for RMSE). MOD out-performance p-value is the combined p-value of MOD NLL being less than the NLL of the method in the corresponding row. \textbf{Bold} indicates best in category and \textit{\textbf{bold+italicized}} indicates second best. In case of a tie in the means, the method with lower standard deviation is highlighted.}
\label{table_dna}
\begin{tabular}{lcccccc}
\toprule
 &  &  \multicolumn{2}{c}{(MOD out-performance)} &  &   \multicolumn{2}{c}{(MOD out-performance)}\\
Methods   &  Out-of-distribution NLL   &   \# of TFs   &   p-value  &   In-distribution NLL   &   \# of TFs  &   p-value \\
\midrule
%\multicolumn{7}{l}{\textbf{In-distribution test performance} }\\
\multicolumn{7}{l}{\textbf{(OOD as sequences with top 10\% binding affinity)} }\\
\methodnormal{}& 0.7485$\pm$0.124 & 26 & 1.7e-05 & -0.4266$\pm$0.031 & 32 & 7.7e-09\\
\methodnormaladv{}& 0.7438$\pm$0.122 & 25 & 0.001 & -0.4312$\pm$0.033 & 26 & 0.005\\
\methodnc{}& 0.7358$\pm$0.118 & 27 & 0.061 & -0.4314$\pm$0.032 & 17 & 0.761\\
\methodabbrev{}& \textbf{0.7153$\pm$0.117} & $-$ & $-$ & -0.4312$\pm$0.031 & $-$ & $-$\\
\idabbrev{}& \textbf{\textit{0.7225$\pm$0.116}} & 22 & 0.359 & \textbf{-0.4325$\pm$0.032} & 16 & 0.777\\
\methodtrain{}& 0.7326$\pm$0.121 & 26 & 0.012 & \textbf{\textit{-0.4317$\pm$0.032}} & 19 & 0.535\\
\midrule
\multicolumn{7}{l}{\textbf{(OOD as sequences with $>$80\% GC content)}}\\
\methodnormal{}& -0.6938$\pm$0.052 & 20 & 0.022 & -0.5649$\pm$0.029 & 34 & 3.1e-11\\
\methodnormaladv{}& \textbf{-0.7010$\pm$0.041} & 23 & 0.007 & \textbf{-0.5740$\pm$0.027} & 21 & 0.292\\
\methodnc{}& -0.6805$\pm$0.065 & 25 & 0.011 & -0.5700$\pm$0.026 & 25 & 0.017\\
\methodabbrev{}& \textbf{\textit{-0.7007$\pm$0.047}} & $-$ & $-$ & \textbf{\textit{-0.5729$\pm$0.027}} & $-$ & $-$ \\
\idabbrev{}& -0.6959$\pm$0.040 & 24 & 0.004 & -0.5720$\pm$0.027 & 22 & 0.357\\
\methodtrain{}& -0.6948$\pm$0.054 & 21 & 0.103 & -0.5711$\pm$0.028 & 22 & 0.163\\
\bottomrule
\end{tabular}
\end{table*}

\begin{figure}[h!]
\centering
  \includegraphics[width=0.9\linewidth,trim=0.5cm 0.5cm 0.5cm 0.5cm, clip=true]{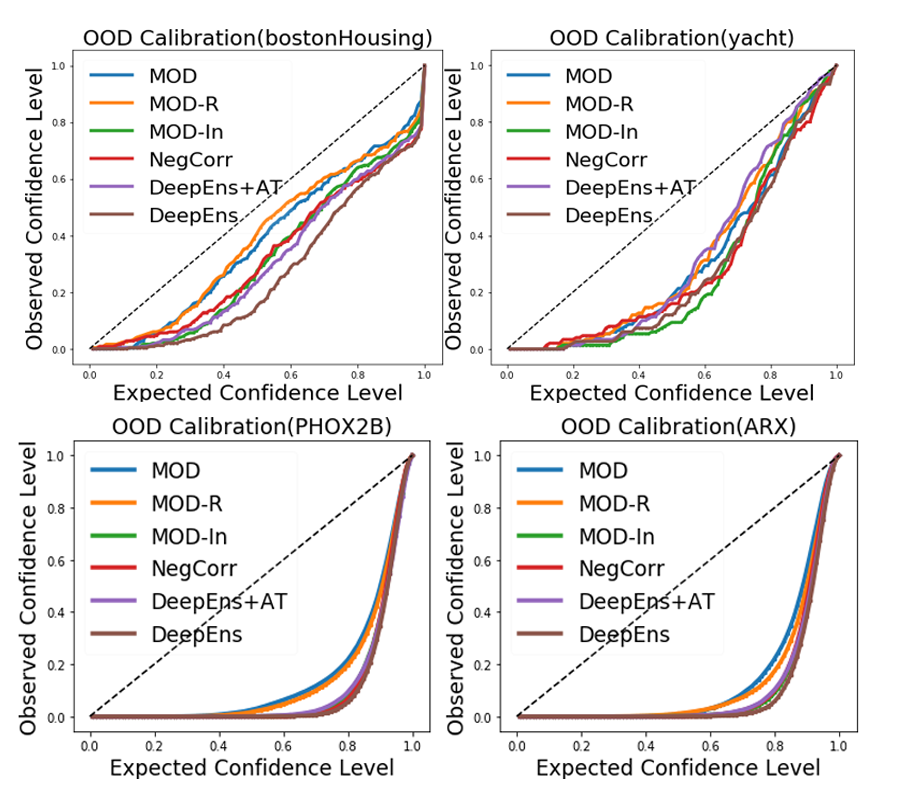}
\caption{Calibration curves for regression models trained on two of the UCI datasets (top) and two DNA TF binding datasets (bottom). A perfect calibration curve should lie on the diagonal, and an over-confident model has calibration curve where the model expected confidence level is higher than observed confidence level (below diagonal). MOD-R and MOD significantly improve the over-confident predictions from the deep ensembles trained without augmentation loss.}
\label{fig:calcurve}
\end{figure}

Table~\ref{table_dna} and Appendix Table 1 shows mean OOD and in-distribution performance across 38 TFs (averaged over 10 runs using random data splits and NN initializations). \methodabbrev{} methods have significantly improved performance on all metrics and OOD setups compared to \methodnormal{}/DeepEns+AT, both in terms of \# of TF outperforming and overall p-value and is on par with \methodnormaladv{} on in-Distribution.  The re-weighting scheme (\idabbrev{}) further improved the performance on top 10\% $Y$-value OOD set up. Figure~\ref{fig:calcurve} shows the calibration curve on two of the TFs where the deep ensembles are over-confident on top 10\% $Y$-value OOD examples. MOD-R and MOD improve the calibration results by significant margin compared to most of the baselines.

% \begin{figure}[!htb]
% \begin{floatrow}
\paragraph{Bayesian Optimization } Next, we compared how the \methodabbrev{}, \idabbrev{}, and \methodtrain{} ensembles performed against the \methodnormal{}, \methodnormaladv{}, and \methodnc{} ensembles in 38 Bayesian optimization tasks using the same protein binding data \cite{Hashimoto18}. For each TF, we performed 30 rounds of DNA-sequence acquisition, acquiring batches of 10 sequences per round in an attempt to maximize binding affinity. We used the \emph{upper confidence bound} (UCB) as our acquisition function \cite{Chen17}, ordering the candidate points via $\widebar{\mu}(x) + \beta \cdot \sigma_{\text{mod}}(x)$ (with UCB coefficient $\beta = 1$).  

\begin{figure}[!htb]
\centering
\begin{minipage}{0.4\textwidth}
  \includegraphics[width=1\linewidth]{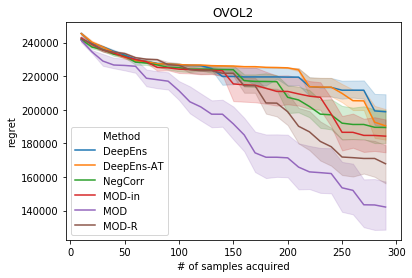}
\end{minipage}
\begin{minipage}{0.4\textwidth}
  \includegraphics[width=1\linewidth]{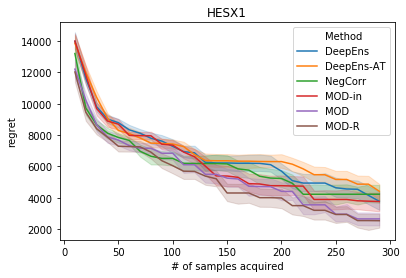}
\end{minipage}

  \caption{Regret for two Bayesian optimization tasks (averaged over 20 replicate runs). The bands depict 50\% confidence intervals, and the $x$-axis indicates the number of DNA sequences whose binding affinity has been profiled so far.}
  \label{fig:regret}

\end{figure}

\begin{table}[!htb]
\small 
\centering\caption{Regret ($r_T$) comparison. Each cell shows the number of TFs (out of 38) for which the method in corresponding row outperforms the method in the corresponding column (lower $r_T$). The number in parentheses is the combined (across 38 TFs) p-value of MOD/-in/-R regret being less than the regret of the method in the corresponding column.}
\label{table:bo_matrix}
\begin{tabular}{lccc}
\toprule
vs & \methodnormal{} & \methodnormaladv & \methodnc{} \\
\midrule
\methodtrain{} & 21 (0.111) & 21 (0.041) & 19 (0.356) \\ 
\methodabbrev{} & 26 (0.003) & 24 (0.004) & 20 (0.001) \\ 
\idabbrev{} & 22 (0.019) & 23 (0.007) & 22 (0.017) \\ 
\bottomrule
vs & \methodtrain{} & \methodabbrev{} & \idabbrev{} \\
\midrule
\methodtrain{} & $-$ & 17 (0.791) & 16 (0.51) \\ 
\methodabbrev{} & 19 (0.002) &  $-$ & 22 (0.173) \\ 
\idabbrev{} & 20 (0.052) & 14 (0.674) &  $-$ \\ 
\bottomrule
\end{tabular}
\end{table}
At every acquisition iteration, we randomly held out 10\% of the training set as the validation set and chose the  $\gamma$ penalty (for \methodabbrev{}, \methodtrain{}, \idabbrev{}, and \methodnc{}) that produced the best validation NLL (out of choices: 0, 5, 10, 20, 40, 80). The stopping epoch is chosen based on the validation NLL not increasing for 10 epochs with an upper limit of 30 epochs. Optimization was done with a learning rate of 0.01, L2 penalty of 0.01 and used the Adam optimizer. For each of the 38 TFs, we performed 20 Bayesian optimization runs with different seed sequences (same seeds used for all the methods) and using 200 points randomly sampled from the bottom 90\% of $Y$ values as are initial training set.

We evaluated on the metric of simple regret $r_T = \max_{x\in \mathcal{X}} f(x) - \max_{t\in [1,T]} f(x_t)$ (second term in the subtraction quantifies the best point acquired so far and the first term is the global best). The results are presented in Table~\ref{table:bo_matrix}. \methodabbrev{} outperforms all other methods in both number of TFs with better regret and the combined p-value. \idabbrev{} is also strong outperforming all other methods except \methodabbrev{} with respect to which is about equivalent in terms of statistical significance. Figure~\ref{fig:regret} shows $r_T$ for the TFs \emph{OVOL2} and \emph{HESX1},  a task in which \methodabbrev{} and \idabbrev{} outperform the other methods.

\begin{comment}
\begin{table}[!htb]
\centering\caption{\textbf{Pairwise  $r_T$ comparison} The first number of each cell in the matrix is the number of TFs on which the method for that row outperformed (in terms of $r_T$) the method for that column with a p-value of $\leq 0.05$ (computed using a one-tailed paired t-test). The second number (in parenthesis) of the cell is the combined p-value (combined using Fisher's method) of the row method outperforming the column method. Bolded counts/p-values indicate superior performance.}

\label{table:bo_matrix}
\begin{tabular}{|c|p{13mm}|p{14mm}|p{15mm}|p{15mm}|p{13mm}|p{13mm}|}
\hline
X & \textbf{\methodnormal{}} & \textbf{\methodnormaladv{}} & \textbf{\methodnc{}} & \textbf{\methodtrain{}} &  \textbf{\methodabbrev{}} & \textbf{\idabbrev{}} \\
\hline 
\textbf{\methodnormal{}} & X & \bf{2 (0.028)} & 0 & 1 (0.069) & 1 (0.31) & 1 (0.181) \\
\textbf{\methodnormaladv{}} & 2 (0.049) & X & \bf{3 (0.005)} & \bf{6 (1E-4)} & 1 (0.35) & 2 (0.053) \\
\textbf{\methodnc{}} & \bf{1 (0.036)} & 3 (0.016) & X & 1 (0.066) & 0 & 2 (0.032) \\
\textbf{\methodtrain{}} & \bf{2 (0.005)} & 3 (0.019) & \bf{1 (0.031)} & X & 2 (0.145) & 2 (0.03) \\
\textbf{\methodabbrev{}} & \bf{5 (1E-4)} & \bf{6 (9E-5)} & \bf{6 (<1E-6)} & \bf{8 (<1E-6)} & X & 1 (0.064) \\
\textbf{\idabbrev{}} & \bf{3 (7E-4)} & \bf{5 (4E-4)} & \bf{3 (0.002)} & \bf{4 (0.001)} & \textbf{2 (0.028)} & X \\
\hline
\end{tabular}
\end{table}
\end{comment}

%\todo[comment on practical difference between variance penalty vs.\ correlation/diversity penalty]

\subsection{UCI Regression Datasets}
We next experimented with 9 real world datasets with continuous inputs in some applicable bounded domain. We follow the experimental setup that~\cite{Lakshminarayanan17} and~\cite{hernandez2017parallel} used to evaluate deep ensembles and deep Bayesian regressors. % probabilistic back-propagation. 
We split off all datapoints whose $y$-values fall in the top $5\%$ as an OOD test set (so datapoints with such large $y$-values are never encountered during training). We simulate the situation where training set is limited and thus used $40\%$ of the data for training and $10\%$ for validation. The remaining data is used as an in-distribution test set. The analysis is repeated for 10 random splits of the data to ensure robustness. We again use an ensemble of 4 fully-connected neural networks with the same architecture as above and the NLL training loss searching over hyperparameter values: L2 penalty  $\in \{0,0.001,0.01,0.05,0.1\}$,  learning rate $ \in \{ 0.0005,0.001, 0.01\}$. We report the negative log-likelihood (NLL) on both in- and out-of-distribution test sets for ensembles trained via different strategies (including \methodadv{}) and examine the calibration curves.

As shown in Table~\ref{table_uci}, \methodabbrev{} outperforms \methodnormal{} in 6 out of the 9 datasets on OOD NLL, and has significant overall p-value compared to all baselines. \methodadv{} ranks top 1 in OOD NLL in terms of averaged ranks across all datasets, showing better robustness than \methodabbrev{}. The MOD loss lead to higher-quality uncertainties on OOD data while also improving in-distribution performance of \methodnormal{}.

\begin{table*}[h]
%\normal 
\centering\caption{Averaged NLL on out-of-distribution/in-distribution test example over 10 replicate runs for UCI datasets, top 5\% samples were heldout as OOD test set (See Appendix Table 2 for RMSE). MOD outperformance p-value is the combined (via Fisher's method) p-value of MOD NLL being less than the NLL of the method in the corresponding column (with p-value per dataset being computed using a paired single tailed t-test). \textbf{Bold} indicates best in category and \textit{\textbf{bold+italicized}} indicates second best. In case of a tie in the means, the method with the lower standard deviation is highlighted.}
\label{table_uci}

\resizebox{2.1\columnwidth}{!}{
\begin{tabular}{lccccccc}
\toprule
Datasets & \methodnormal{} & \methodnormaladv{} & \methodnc{} & \methodabbrev{} & \idabbrev{} & \methodtrain{} &\methodadv{}\\
\midrule
\multicolumn{8}{l}{\textbf{Out-of-distribution NLL}}\\
concrete& -0.831$\pm$0.237 & -0.915$\pm$0.204 & -0.913$\pm$0.277 & -0.904$\pm$0.118 & -0.910$\pm$0.193 & \textbf{\textit{-0.924$\pm$0.188}} & \textbf{-0.950$\pm$0.200}\\
yacht& -1.597$\pm$0.840 & -1.762$\pm$0.647 & \textbf{-1.972$\pm$0.570} & -1.797$\pm$0.437 & -1.761$\pm$0.578 & -1.638$\pm$0.663 & \textbf{\textit{-1.948$\pm$0.343}}\\
naval-propulsion-plant& -2.580$\pm$0.103 & -1.380$\pm$0.087 & -2.618$\pm$0.056 & \textbf{-2.729$\pm$0.071} & -2.130$\pm$0.069 & -2.057$\pm$0.055 & \textbf{\textit{-2.629$\pm$0.068}}\\
wine-quality-red& 0.133$\pm$0.132 & 0.115$\pm$0.086 & 0.113$\pm$0.104 & 0.153$\pm$0.107 & \textbf{0.084$\pm$0.072} & \textbf{\textit{0.085$\pm$0.065}} & 0.217$\pm$0.114\\
power-plant& \textbf{-1.734$\pm$0.054} & -1.731$\pm$0.088 & -1.659$\pm$0.075 & -1.638$\pm$0.151 & -1.644$\pm$0.120 & \textbf{\textit{-1.731$\pm$0.050}} & -1.669$\pm$0.066\\
protein-tertiary-structure& \textbf{\textit{1.162$\pm$0.231}} & 1.178$\pm$0.158 & 1.231$\pm$0.130 & 1.197$\pm$0.137 & 1.194$\pm$0.214 & \textbf{1.154$\pm$0.132} & 1.299$\pm$0.252\\
kin8nm& -1.980$\pm$0.053 & -1.970$\pm$0.093 & \textbf{-2.036$\pm$0.046} & -1.999$\pm$0.049 & -2.003$\pm$0.095 & -1.993$\pm$0.078 & \textbf{\textit{-2.027$\pm$0.085}}\\
bostonHousing& 1.591$\pm$0.680 & 1.243$\pm$0.690 & 1.821$\pm$0.913 & \textbf{\textit{0.568$\pm$0.959}} & \textbf{0.460$\pm$0.648} & 0.923$\pm$0.733 & 1.517$\pm$0.711\\
energy& -1.590$\pm$0.253 & \textbf{-1.784$\pm$0.153} & -1.718$\pm$0.193 & -1.736$\pm$0.117 & -1.741$\pm$0.264 & -1.733$\pm$0.199 & \textbf{\textit{-1.772$\pm$0.242}}\\
MOD outperformance p-value& 0.002 & 4.9e-07 & 0.034 & - & 1.6e-04 & 4.6e-05 & 0.027\\
\midrule
\multicolumn{8}{l}{\textbf{In-distribution NLL}}\\
concrete& -1.075$\pm$0.094 & -1.129$\pm$0.084 & -1.089$\pm$0.102 & \textbf{-1.155$\pm$0.086} & \textbf{\textit{-1.137$\pm$0.132}} & -1.090$\pm$0.092 & -1.047$\pm$0.177\\
yacht& -3.286$\pm$0.692 & -3.245$\pm$0.822 & \textbf{-3.570$\pm$0.166} & -3.500$\pm$0.190 & -3.461$\pm$0.252 & -3.339$\pm$0.815 & \textbf{\textit{-3.556$\pm$0.203}}\\
naval-propulsion-plant& -2.735$\pm$0.077 & -1.513$\pm$0.042 & -2.810$\pm$0.042 & \textbf{-2.857$\pm$0.067} & -2.297$\pm$0.061 & -2.238$\pm$0.047 & \textbf{\textit{-2.817$\pm$0.046}}\\
wine-quality-red& -0.070$\pm$0.853 & -0.341$\pm$0.068 & -0.266$\pm$0.291 & -0.337$\pm$0.069 & \textbf{\textit{-0.348$\pm$0.045}} & \textbf{-0.351$\pm$0.055} & -0.170$\pm$0.505\\
power-plant& -1.521$\pm$0.015 & \textbf{-1.525$\pm$0.018} & -1.524$\pm$0.023 & -1.523$\pm$0.012 & -1.522$\pm$0.017 & \textbf{\textit{-1.524$\pm$0.013}} & -1.523$\pm$0.016\\
protein-tertiary-structure& -0.514$\pm$0.013 & -0.519$\pm$0.007 & \textbf{-0.544$\pm$0.012} & -0.533$\pm$0.009 & -0.532$\pm$0.012 & -0.529$\pm$0.008 & \textbf{\textit{-0.540$\pm$0.012}}\\
kin8nm& -1.305$\pm$0.016 & -1.315$\pm$0.020 & \textbf{-1.334$\pm$0.015} & -1.317$\pm$0.019 & -1.315$\pm$0.017 & \textbf{\textit{-1.322$\pm$0.015}} & -1.314$\pm$0.020\\
bostonHousing& -0.901$\pm$0.154 & \textbf{\textit{-0.937$\pm$0.144}} & -0.656$\pm$0.671 & \textbf{-0.953$\pm$0.147} & -0.883$\pm$0.188 & -0.925$\pm$0.180 & -0.728$\pm$0.376\\
energy& -2.426$\pm$0.151 & -2.517$\pm$0.098 & \textbf{\textit{-2.620$\pm$0.130}} & -2.507$\pm$0.153 & -2.525$\pm$0.098 & -2.522$\pm$0.129 & \textbf{-2.638$\pm$0.137}\\
MOD outperformance p-value& 2.4e-08 & 6.9e-11 & 0.116 & - & 8.9e-06 & 2.2e-06 & 0.046\\
\bottomrule
\end{tabular}
}
\end{table*}

Figure~\ref{fig:calcurve} shows the calibration curve on two of the datasets where the basic deep ensembles exhibit over-confidence on OOD data. Note that retaining accurate calibration on OOD data is extremely difficult for most machine learning methods. 
MOD and MOD-R improve calibration by a significant margin compared to most of the baselines, validating the effectiveness of our MOD procedure.

% We also show the mean $r_T$ (Appendix Figure~\ref{fig:mean_regret}) across the 38 TFs $\times$ 10 BO samples per TF. 
%On most tasks, \idabbrev{}-BO, and \methodabbrev{}-BO typically outperform \met-BO, and \methodnormal{}-BO in the later acquisition rounds and are also able to achieve lower average regret over all 38 tasks.

\begin{figure}[h!]
\centering
  \includegraphics[width=0.95\linewidth,trim=0 0 0 0, clip=true]{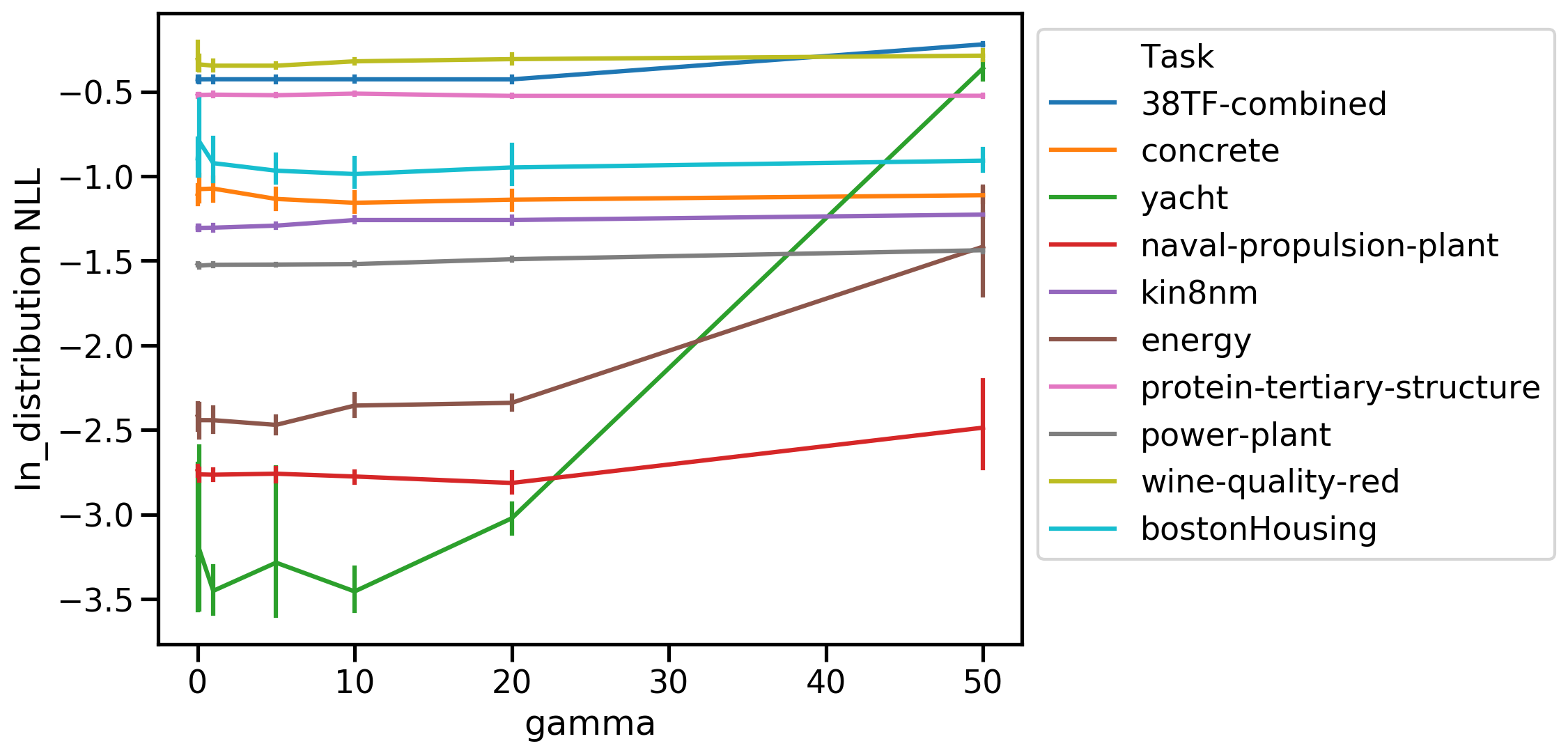}
\caption{The effect of different $\gamma$ on in-distribution test performance (NLL).}
\label{fig:gamma}
\end{figure}

The selection of $\gamma$ is critical for \methodabbrev{}, thus we also examine the effect of the choice of $\gamma$ on the in-distribution performance for the 9 UCI and 38 TF binding regression tasks. As shown in Figure~\ref{fig:gamma}, $\gamma$
generally does not affect or hurt in-distribution NLL until it gets too large at which point it fairly consistently starts hurting it. When $\gamma$ is selected properly it may even improve the in-distribution slightly as shown in the previous tables.

\subsection{Age Prediction from Images}

To demonstrate the effectiveness of \methodabbrev{}, \methodadv{}, and \idabbrev{} on high dimensional data, we consider supervised learning with image data. Here, we use a dataset of human images collected from IMDB and Wikimedia and annotated with age and gender information~\cite{rothe2015dex}. The IMDB/Wiki parts of the dataset consist of 460K+/62K+ images respectively. 28,601 images in the Wiki dataset are males and the rest are females.

In the context of Wiki images, we tried to predict the ages given the image of a person using 2000 images of \emph{males as the training set}. For the OOD dataset, we hold out the oldest 10\% of the people as the OOD set.
We used the Wide Residual Network architecture~\cite{zagoruyko2016wide} with a depth of 4 and a width factor of 2. As before, we used an ensemble of size 4. The search for the optimal $\gamma$ value was over $0, 2, 5, 10, 20, 40, 80$. The stopping epoch is chosen based on the validation NLL not increasing for 10 epochs with an upper limit of 30 epochs. Optimization was done with a learning rate of 0.001, l2 penalty of 0.001 and used the Adam optimizer. The NLL results are in Table~\ref{table:image_reg_nll} whereas the RMSE results are in the Appendix. Both \methodname{} and \methodadv{} get the best results on OOD NLL with the improvement being statistically significant over the other methods. \methodabbrev{} gets an NLL of 1.129 on OOD data, \methodadv{} gets an NLL of 1.155 on OOD, and  \idabbrev{} gets 1.185 on OOD. This is in contrast to \methodnormal{} which gets only 1.304 on OOD.  Thus both \methodabbrev{} and \idabbrev{} show significant improvements on NLL on the OOD data. In addition, while \methodnormaladv{} has a better mean {\em in-distribution} NLL compared to \methodabbrev{}, the focus of this paper is out of distribution uncertainty on which \methodname{} and \methodadv{} perform very well. Notably every MOD variant \emph{ improves} performance for {\em both} in and out of distribution. Thus augmenting the loss function with the MOD penalty should not make your model worse.

\begin{table}[!htb]
\centering\caption{Image regression results showing mean performance across 20 randomly seeded runs (along with $\pm$ one standard deviation). In-Dist refers to the in-distribution test set. OOD refers to the out of distribution test set. \textbf{Bold} indicates best in category and \textit{\textbf{bold+italicized}} indicates second best. In case of a tie in means, the lower standard deviation method is highlighted.}
\label{table:image_reg_nll}
\begin{tabular}{lcc}
\toprule
Methods & OOD NLL & In-Dist NLL \\
\midrule
\methodnormal{} & 1.3100 $\pm$ 0.2486 & -0.2193 $\pm$ 0.0207 \\
\methodnormaladv{} & 1.2348$\pm$0.1291 & \textbf{-0.2419$\pm$0.0213} \\
\methodnc{} & 1.1731$\pm$0.1978 & -0.2286$\pm$0.0179 \\
\methodtrain{} & 1.2625$\pm$0.1961 & -0.2301$\pm$0.0128 \\
\methodabbrev{} & \textbf{1.1294$\pm$0.1707} & \textbf{\textit{-0.2306$\pm$0.0148}} \\
\idabbrev{} & 1.1847$\pm$0.2442 & -0.2285$\pm$0.0191 \\
\methodadv{} & \textbf{\textit{1.1547$\pm$0.1865}} & -0.2305$\pm$0.0149 \\
\bottomrule
\end{tabular}

\end{table}

\section{Conclusion}

We have introduced a loss function and data augmentation strategy that helps stabilize distribution uncertainty estimates obtained from model ensembling.  Our method increases model uncertainty over the entire input space while simultaneously maintaining predictive performance, which helps mitigate uncertainty collapse that may arise in small model ensembles.
% We further propose a variant of our method which assesses the distance of an augmented sample from the training distribution and aims to ensure higher model uncertainty in regions with low-density under the training data distribution. 
We further proposed two variants of our method.  \idabbrev{} assesses the distance of an augmented sample from the training distribution and aims to ensure higher model uncertainty in regions with low-density, and \methodadv{} uses adversarial optimization to improve model uncertainty on relatively over-confident regions more efficiently. 
Our methods produce improvements to both the in and out of distribution NLL, out of distribution RMSE, and calibration on a variety of datasets drawn from biology, vision, and common UCI datasets. We also showed \methodabbrev{} is useful in hard Bayesian optimization tasks. Future work could develop techniques to generate OOD augmented samples for structured data, as well as applying ensembles with improved uncertainty-awareness to currently challenging tasks such as exploration in reinforcement learning.

%\clearpage
\bibliographystyle{aaai}
\bibliography{mod}

\begin{thebibliography}{}

\bibitem[\protect\citeauthoryear{Balan \bgroup et al\mbox.\egroup
  }{2015}]{darkknowledge}
Balan, A.~K.; Rathod, V.; Murphy, K.~P.; and Welling, M.
\newblock 2015.
\newblock Bayesian dark knowledge.
\newblock In {\em Advances in Neural Information Processing Systems}.

\bibitem[\protect\citeauthoryear{Barrera \bgroup et al\mbox.\egroup
  }{2016}]{barrera2016survey}
Barrera, L.~A.; Vedenko, A.; Kurland, J.~V.; Rogers, J.~M.; Gisselbrecht,
  S.~S.; Rossin, E.~J.; Woodard, J.; Mariani, L.; Kock, K.~H.; Inukai, S.;
  et~al.
\newblock 2016.
\newblock Survey of variation in human transcription factors reveals prevalent
  {DNA} binding changes.
\newblock {\em Science} 351(6280):1450--1454.

\bibitem[\protect\citeauthoryear{Beluch \bgroup et al\mbox.\egroup
  }{2018}]{activelearning}
Beluch, W.~H.; Genewein, T.; N{\"u}rnberger, A.; and K{\"o}hler, J.~M.
\newblock 2018.
\newblock The power of ensembles for active learning in image classification.
\newblock In {\em IEEE Conference on Computer Vision and Pattern Recognition}.

\bibitem[\protect\citeauthoryear{Blundell \bgroup et al\mbox.\egroup
  }{2015}]{blundell2015weight}
Blundell, C.; Cornebise, J.; Kavukcuoglu, K.; and Wierstra, D.
\newblock 2015.
\newblock Weight uncertainty in neural networks.
\newblock {\em arXiv preprint arXiv:1505.05424}.

\bibitem[\protect\citeauthoryear{Breiman}{1996}]{bootstrap}
Breiman, L.
\newblock 1996.
\newblock Bagging predictors.
\newblock {\em Machine Learning} 24:123--140.

\bibitem[\protect\citeauthoryear{Brown}{2004}]{Brown04}
Brown, G.
\newblock 2004.
\newblock {\em Diversity in neural network ensembles}.
\newblock Ph.D. Dissertation, University of Birmingham.

\bibitem[\protect\citeauthoryear{Chen \bgroup et al\mbox.\egroup
  }{2017}]{Chen17}
Chen, R.~Y.; Sidor, S.; Abbeel, P.; and Schulman, J.
\newblock 2017.
\newblock {UCB} exploration via {Q-}ensembles.
\newblock {\em arXiv:1706.01502}.

\bibitem[\protect\citeauthoryear{Choi and Jang}{2018}]{choi2018generative}
Choi, H., and Jang, E.
\newblock 2018.
\newblock Generative ensembles for robust anomaly detection.
\newblock {\em arXiv preprint arXiv:1810.01392}.

\bibitem[\protect\citeauthoryear{Hafner \bgroup et al\mbox.\egroup
  }{2018}]{Hafner18}
Hafner, D.; Tran, D.; Lillicrap, T.; Irpan, A.; and Davidson, J.
\newblock 2018.
\newblock Reliable uncertainty estimates in deep neural networks using noise
  contrastive priors.
\newblock {\em arXiv:1807.09289}.

\bibitem[\protect\citeauthoryear{Hashimoto, Yadlowsky, and
  Duchi}{2018}]{Hashimoto18}
Hashimoto, T.~B.; Yadlowsky, S.; and Duchi, J.~C.
\newblock 2018.
\newblock Derivative free optimization via repeated classification.
\newblock In {\em International Conference on Artificial Intelligence and
  Statistics}.

\bibitem[\protect\citeauthoryear{Hern{\'a}ndez-Lobato \bgroup et
  al\mbox.\egroup }{2017}]{hernandez2017parallel}
Hern{\'a}ndez-Lobato, J.~M.; Requeima, J.; Pyzer-Knapp, E.~O.; and
  Aspuru-Guzik, A.
\newblock 2017.
\newblock Parallel and distributed thompson sampling for large-scale
  accelerated exploration of chemical space.
\newblock {\em arXiv:1706.01825}.

\bibitem[\protect\citeauthoryear{Hooker and Rosset}{2012}]{dar}
Hooker, G., and Rosset, S.
\newblock 2012.
\newblock Prediction-focused regularization using data-augmented regression.
\newblock {\em Statistics and Computing} 1:237--349.

\bibitem[\protect\citeauthoryear{Kuncheva and Whitaker}{2003}]{Kuncheva03}
Kuncheva, L.~I., and Whitaker, C.~J.
\newblock 2003.
\newblock Measures of diversity in classifier ensembles and their relationship
  with the ensemble accuracy.
\newblock {\em Machine Learning} 51:181--207.

\bibitem[\protect\citeauthoryear{Lakshminarayanan, Pritzel, and
  Blundell}{2017}]{Lakshminarayanan17}
Lakshminarayanan, B.; Pritzel, A.; and Blundell, C.
\newblock 2017.
\newblock Simple and scalable predictive uncertainty estimation using deep
  ensembles.
\newblock In {\em Advances in Neural Information Processing Systems}.

\bibitem[\protect\citeauthoryear{Lee \bgroup et al\mbox.\egroup }{2015}]{Lee15}
Lee, S.; Purushwalkam, S.; Cogswell, M.; Crandall, D.; and Batra, D.
\newblock 2015.
\newblock {Why M heads are better than one: Training a diverse ensemble of deep
  networks}.
\newblock {\em arXiv:1511.06314}.

\bibitem[\protect\citeauthoryear{Lee \bgroup et al\mbox.\egroup }{2018}]{Lee18}
Lee, K.; Lee, H.; Lee, K.; and Shin, J.
\newblock 2018.
\newblock Training confidence-calibrated classifiers for detecting
  out-of-distribution samples.
\newblock In {\em International Conference on Learning Representations}.

\bibitem[\protect\citeauthoryear{Liang, Li, and
  Srikant}{2017}]{liang2017enhancing}
Liang, S.; Li, Y.; and Srikant, R.
\newblock 2017.
\newblock Enhancing the reliability of out-of-distribution image detection in
  neural networks.
\newblock {\em arXiv preprint arXiv:1706.02690}.

\bibitem[\protect\citeauthoryear{Liu and Yao}{1999}]{liu1999ensemble}
Liu, Y., and Yao, X.
\newblock 1999.
\newblock Ensemble learning via negative correlation.
\newblock {\em Neural networks} 12(10):1399--1404.

\bibitem[\protect\citeauthoryear{Malinin and
  Gales}{2018}]{malinin2018predictive}
Malinin, A., and Gales, M.
\newblock 2018.
\newblock Predictive uncertainty estimation via prior networks.
\newblock In {\em Advances in Neural Information Processing Systems},
  7047--7058.

\bibitem[\protect\citeauthoryear{Osband \bgroup et al\mbox.\egroup
  }{2016}]{bootstrapdqn}
Osband, I.; Blundell, C.; Pritzel, A.; and Van~Roy, B.
\newblock 2016.
\newblock Deep exploration via bootstrapped {DQN}.
\newblock In {\em Advances in Neural Information Processing Systems}.

\bibitem[\protect\citeauthoryear{Papadopoulos, Edwards, and
  Murray}{2001}]{Papadopoulos01}
Papadopoulos, G.; Edwards, P.~J.; and Murray, A.~F.
\newblock 2001.
\newblock {Confidence estimation methods for neural networks: A practical
  comparison}.
\newblock {\em IEEE Transactions on Neural Networks} 12:1278--1287.

\bibitem[\protect\citeauthoryear{Papernot and
  McDaniel}{2018}]{papernot2018deep}
Papernot, N., and McDaniel, P.
\newblock 2018.
\newblock Deep k-nearest neighbors: Towards confident, interpretable and robust
  deep learning.
\newblock {\em arXiv preprint arXiv:1803.04765}.

\bibitem[\protect\citeauthoryear{Pearce \bgroup et al\mbox.\egroup
  }{2018}]{pearce2018uncertainty}
Pearce, T.; Zaki, M.; Brintrup, A.; and Neel, A.
\newblock 2018.
\newblock Uncertainty in neural networks: Bayesian ensembling.
\newblock {\em arXiv preprint arXiv:1810.05546}.

\bibitem[\protect\citeauthoryear{Ren \bgroup et al\mbox.\egroup
  }{2019}]{ren2019likelihood}
Ren, J.; Liu, P.~J.; Fertig, E.; Snoek, J.; Poplin, R.; DePristo, M.~A.;
  Dillon, J.~V.; and Lakshminarayanan, B.
\newblock 2019.
\newblock Likelihood ratios for out-of-distribution detection.
\newblock {\em arXiv preprint arXiv:1906.02845}.

\bibitem[\protect\citeauthoryear{Riquelme, Tucker, and
  Snoek}{2018}]{deepbandit}
Riquelme, C.; Tucker, G.; and Snoek, J.
\newblock 2018.
\newblock Deep bayesian bandits showdown: An empirical comparison of bayesian
  deep networks for thompson sampling.
\newblock In {\em International Conference on Learning Representations}.

\bibitem[\protect\citeauthoryear{Rothe, Timofte, and
  Van~Gool}{2015}]{rothe2015dex}
Rothe, R.; Timofte, R.; and Van~Gool, L.
\newblock 2015.
\newblock Dex: Deep expectation of apparent age from a single image.
\newblock In {\em Proceedings of the IEEE International Conference on Computer
  Vision Workshops},  10--15.

\bibitem[\protect\citeauthoryear{Shui \bgroup et al\mbox.\egroup
  }{2018}]{Shui18}
Shui, C.; Mozafari, A.~S.; Marek, J.; Hedhli, I.; and Gagne, C.
\newblock 2018.
\newblock Diversity regularization in deep ensembles.
\newblock {\em arXiv:1802.07881}.

\bibitem[\protect\citeauthoryear{Vyas \bgroup et al\mbox.\egroup
  }{2018}]{vyas2018out}
Vyas, A.; Jammalamadaka, N.; Zhu, X.; Das, D.; Kaul, B.; and Willke, T.~L.
\newblock 2018.
\newblock Out-of-distribution detection using an ensemble of self supervised
  leave-out classifiers.
\newblock In {\em Proceedings of the European Conference on Computer Vision
  (ECCV)},  550--564.

\bibitem[\protect\citeauthoryear{Zagoruyko and
  Komodakis}{2016}]{zagoruyko2016wide}
Zagoruyko, S., and Komodakis, N.
\newblock 2016.
\newblock Wide residual networks.
\newblock {\em arXiv preprint arXiv:1605.07146}.

\end{thebibliography}
\onecolumn
\newpage

\section*{Appendix}

\begin{table*}[h]
\centering\caption{RMSE on OOD/in-distribution test set averaged across 38 TFs over 10 replicate runs. MOD out-performance p-value is the combined p-value of MOD RMSE being less than the RMSE of the method in the corresponding row. \textbf{Bold} indicates best in category and \textit{\textbf{bold+italicized}} indicates second best. In case of a tie in the means, the method with lower standard deviation is highlighted.}
\label{table_dna_rmse}
\begin{tabular}{lcccccc}
\toprule
 &  &  \multicolumn{2}{c}{(MOD out-performance)} &  &   \multicolumn{2}{c}{(MOD out-performance)}\\
Methods   &  Out-of-distribution RMSE   &   \# of TFs   &   p-value  &   In-distribution RMSE   &   \# of TFs  &   p-value \\
\midrule
\multicolumn{7}{l}{\textbf{(OOD as sequences with top 10\% binding affinity)} }\\
\methodnormal{}& 0.2837$\pm$0.011 & 26 & 1.3e-04 & 0.1591$\pm$0.005 & 34 & 2.1e-12\\
\methodnormaladv{}& 0.2812$\pm$0.011 & 23 & 0.087 & 0.1582$\pm$0.005 & 28 & 3.6e-05\\
\methodnc{}& 0.2814$\pm$0.010 & 20 & 0.124 & 0.1583$\pm$0.005 & 20 & 0.492\\
\methodabbrev{}& 0.2802$\pm$0.010 & 0 & 0.0e+00 & \textbf{\textit{0.1581$\pm$0.005}} & 0 & 0.0e+00\\
\idabbrev{}& \textbf{0.2795$\pm$0.010} & 17 & 0.933 & \textbf{0.1579$\pm$0.005} & 13 & 0.969\\
\methodtrain{}& \textbf{\textit{0.2801$\pm$0.010}} & 16 & 0.617 & 0.1581$\pm$0.005 & 18 & 0.633\\
\midrule
\multicolumn{7}{l}{\textbf{(OOD as sequences with $>$80\% GC content)}}\\
\methodnormal{}& 0.1190$\pm$0.004 & 25 & 0.008 & 0.1415$\pm$0.003 & 36 & 8.5e-24\\
\methodnormaladv{}& 0.1180$\pm$0.003 & 19 & 0.029 & \textbf{\textit{0.1394$\pm$0.003}} & 22 & 0.106\\
\methodnc{}& 0.1179$\pm$0.004 & 21 & 0.052 & 0.1403$\pm$0.003 & 30 & 4.7e-08\\
\methodabbrev{}& \textbf{0.1173$\pm$0.003} & 0 & 0.0e+00 & \textbf{0.1394$\pm$0.003} & 0 & 0.0e+00\\
\idabbrev{}& \textbf{\textit{0.1177$\pm$0.003}} & 22 & 0.112 & 0.1398$\pm$0.003 & 19 & 0.079\\
\methodtrain{}& 0.1177$\pm$0.004 & 21 & 0.401 & 0.1403$\pm$0.003 & 27 & 2.8e-06\\
\bottomrule
\end{tabular}
\end{table*}

\begin{table*}[h]
\label{table_regression-rmse}
\caption{RMSE on out-of-distribution/in-distribution test examples for the UCI datasets (over 10 replicate runs). Examples with $Y$-values in the top 5\% were held out as the OOD test set.}
\resizebox{1\columnwidth}{!}{
\begin{tabular}{lccccccc}
\toprule
Datasets & \methodnormal{} & \methodnormaladv{} & \methodnc{} & \methodabbrev{} & \idabbrev{} & \methodtrain{} & \methodadv{}\\
\midrule
\multicolumn{8}{l}{\textbf{Out-of-distribution RMSE}}\\
concrete& 0.105$\pm$0.016 & 0.102$\pm$0.011 & \textbf{\textit{0.096$\pm$0.012}} & 0.105$\pm$0.014 & 0.113$\pm$0.028 & 0.099$\pm$0.011 & \textbf{0.095$\pm$0.012}\\
yacht& 0.038$\pm$0.041 & 0.039$\pm$0.044 & \textbf{0.018$\pm$0.005} & 0.026$\pm$0.005 & 0.026$\pm$0.007 & 0.039$\pm$0.045 & \textbf{\textit{0.021$\pm$0.005}}\\
naval-propulsion-plant& \textbf{\textit{0.111$\pm$0.008}} & 0.126$\pm$0.015 & 0.112$\pm$0.011 & \textbf{0.109$\pm$0.008} & 0.125$\pm$0.008 & 0.129$\pm$0.007 & 0.122$\pm$0.009\\
wine-quality-red& 0.239$\pm$0.013 & 0.236$\pm$0.009 & \textbf{\textit{0.234$\pm$0.008}} & 0.239$\pm$0.011 & \textbf{0.234$\pm$0.010} & 0.235$\pm$0.008 & 0.242$\pm$0.014\\
power-plant& \textbf{\textit{0.041$\pm$0.002}} & 0.042$\pm$0.004 & 0.044$\pm$0.004 & 0.048$\pm$0.012 & 0.047$\pm$0.010 & \textbf{0.041$\pm$0.003} & 0.043$\pm$0.003\\
protein-tertiary-structure& 0.301$\pm$0.009 & 0.301$\pm$0.003 & \textbf{0.294$\pm$0.003} & \textbf{\textit{0.299$\pm$0.008}} & 0.300$\pm$0.011 & 0.300$\pm$0.007 & 0.300$\pm$0.004\\
kin8nm& 0.042$\pm$0.005 & 0.040$\pm$0.005 & \textbf{0.037$\pm$0.003} & \textbf{\textit{0.039$\pm$0.003}} & 0.041$\pm$0.005 & 0.039$\pm$0.004 & 0.039$\pm$0.004\\
bostonHousing& 0.221$\pm$0.023 & 0.213$\pm$0.021 & 0.210$\pm$0.016 & \textbf{\textit{0.209$\pm$0.017}} & \textbf{0.200$\pm$0.014} & 0.212$\pm$0.016 & 0.217$\pm$0.022\\
energy& 0.060$\pm$0.014 & 0.054$\pm$0.012 & 0.053$\pm$0.009 & \textbf{\textit{0.053$\pm$0.005}} & 0.054$\pm$0.013 & 0.054$\pm$0.009 & \textbf{0.047$\pm$0.009}\\
MOD outperformance p-value& 0.028 & 0.102 & 0.989 & 0.0e+00 & 0.090 & 0.017 & 0.124\\
\midrule
\multicolumn{8}{l}{\textbf{In-distribution RMSE}}\\
concrete& 0.086$\pm$0.004 & 0.085$\pm$0.004 & 0.084$\pm$0.004 & \textbf{\textit{0.083$\pm$0.003}} & \textbf{0.083$\pm$0.004} & 0.084$\pm$0.004 & 0.084$\pm$0.004\\
yacht& 0.017$\pm$0.019 & 0.019$\pm$0.025 & \textbf{0.010$\pm$0.002} & 0.012$\pm$0.003 & 0.012$\pm$0.004 & 0.018$\pm$0.023 & \textbf{\textit{0.010$\pm$0.002}}\\
naval-propulsion-plant& 0.080$\pm$0.003 & 0.089$\pm$0.003 & \textbf{\textit{0.079$\pm$0.004}} & \textbf{0.077$\pm$0.003} & 0.088$\pm$0.003 & 0.091$\pm$0.004 & 0.085$\pm$0.004\\
wine-quality-red& 0.170$\pm$0.005 & 0.170$\pm$0.005 & 0.169$\pm$0.004 & 0.170$\pm$0.003 & \textbf{0.169$\pm$0.004} & \textbf{\textit{0.169$\pm$0.005}} & 0.170$\pm$0.004\\
power-plant& 0.053$\pm$0.001 & 0.053$\pm$0.001 & \textbf{0.052$\pm$0.001} & 0.053$\pm$0.001 & 0.053$\pm$0.001 & 0.053$\pm$0.001 & \textbf{\textit{0.052$\pm$0.001}}\\
protein-tertiary-structure& 0.164$\pm$0.002 & 0.164$\pm$0.001 & \textbf{0.161$\pm$0.001} & \textbf{\textit{0.163$\pm$0.000}} & 0.163$\pm$0.001 & 0.163$\pm$0.001 & 0.164$\pm$0.001\\
kin8nm& 0.074$\pm$0.002 & 0.072$\pm$0.003 & \textbf{0.071$\pm$0.002} & 0.072$\pm$0.002 & 0.073$\pm$0.001 & \textbf{\textit{0.072$\pm$0.001}} & 0.073$\pm$0.002\\
bostonHousing& 0.085$\pm$0.008 & \textbf{\textit{0.084$\pm$0.008}} & \textbf{0.083$\pm$0.007} & 0.084$\pm$0.009 & 0.084$\pm$0.008 & 0.084$\pm$0.009 & 0.085$\pm$0.009\\
energy& 0.042$\pm$0.004 & 0.039$\pm$0.003 & \textbf{0.036$\pm$0.004} & 0.039$\pm$0.004 & 0.039$\pm$0.003 & 0.039$\pm$0.003 & \textbf{\textit{0.037$\pm$0.004}}\\
In-dist RMSEpval Outperformed by MOD& 4.0e-06 & 5.9e-07 & 0.943 & 0.0e+00 & 8.9e-04 & 0.002 & 0.002\\
\bottomrule
\end{tabular}
}
\end{table*}

\begin{table*}[h]
%\small
\centering
\caption{Image regression results showing mean performance across 20 randomly seeded runs (along with $\pm$ one standard deviation). In-Dist refers to the in-distribution test set. OOD refers to the out of distribution test set. \textbf{Bold} indicates best in category and \textit{\textbf{bold+italicized}} indicates second best. In case of a tie in means, the lower standard deviation method is highlighted.}
\label{table:image_reg_rmse}
\begin{tabular}{lcc}
\toprule
Methods & OOD RMSE & In-Dist RMSE \\
\midrule
\methodnormal{} & 0.388 $\pm$ 0.021 & 0.196 $\pm$ 0.003\\
\methodnormaladv{} & 0.378 $\pm$0.016 & \textbf{0.192$\pm$0.004} \\
\methodnc{} & 0.376$\pm$ 0.015 & 0.194$\pm$ 0.002\\
\methodtrain{} & 0.382$\pm$0.016 & 0.194$\pm$0.002 \\
\methodabbrev{} & \textbf{\textit{0.375$\pm$0.014}} & 0.194$\pm$0.002\\
\idabbrev{} & 0.377$\pm$0.019 & 0.193$\pm$0.002 \\
\methodadv{} & \textbf{0.374$\pm$0.016} & 0.193$\pm$0.002 \\
\bottomrule
\end{tabular}
\end{table*}

% \subsection{References} 

% \begin{quote}
% \begin{small}
% \bibliographystyle{aaai}
% \bibliography{mod}
% \end{small}
% \end{quote}

\end{document}